# An Internal Arc Fixation Channel and Automatic Planning Algorithm for Pelvic Fracture

Qing Yang[1], Jian Song[1], Chang Cheng[2], Chao Shi[1], Chendi Liang[1], Yu Wang[1,*]

[1]*School of Biological Science and Medical Engineering, Beihang University, Beijing, China*

yangq@buaa.edu.cn, sjkxkl@163.com, 13361435056@163.com, liangchendi1997@sina.com, wangyu@buaa.edu.cn.

[2] *Department of Mathematics and Computer Science, Colorado College, Colorado, USA*

d_cheng@coloradocollege.edu.

*Abstract*—**Fixating fractured pelvis fragments with the sacroiliac screw is a common treatment for unstable pelvis fracture. Due to the complex shape of the pelvis, sometimes a suitable straight screw fixation channel cannot be found using traditional methods, which increases the difficulty of pelvic fracture fixation. Therefore, there is an urgent need to find a new screw fixation method to improve the feasibility of pelvic fracture fixation. In this study, a new method of arc nail fixation is proposed to treat the pelvic fracture. An algorithm is proposed to verify the feasibility of the internal arc fixation channel (IAFC) in the pelvis, and the algorithm can calculate a relatively optimal IAFC in the pelvis. Furthermore, we compared the advantages and disadvantages of arc channel and straight channel through experiments. This study verified the feasibility of the IAFC, and the comparison of experimental results shows that the adaptability and safety of the arc channel fixation is better than the traditional straight sacroiliac screw.**

## I. INTRODUCTION

Pelvis fracture is a serious trauma with disability rate as high as 60%. The main characteristics of pelvic fracture are pelvic ring disruption and fracture displacement [1]. In the treatment of pelvic fracture, a common method of pelvic ring fixation is to place cannulated screws into intracortical spaces known as osseous fixation paths (OFPs). Commonly, screws are used to fill the available OFPs and thereby stabilize pelvic and acetabular fractures either percutaneously or after a formal open approach [2]. In the current research, many scholars have studied the best fixation position of straight sacroiliac screws [3-5]. However, due to the pelvis' complex shape and the nerve distribution around the pelvis, proper straight screw paths cannot always be found [6-8]. Even when a straight path can be found, doctors often need to try many times to position the straight screw into the appropriate position during operation [9, 10]. These attempts increase the operation time and radiation exposure of patients and clinicians. To address various operational difficulties mentioned above, some scholars have concentrated on the study robot-assisted navigated drilling of bones for pedicle screw placement [11-13]. Zakariaee et al. put forward the idea of arc nail fixation, and showed that viable arc channels for pelvis fracture fixation exist. Their results showed that there are sizable tunnels in which to put large diameter internal fixation of a curved nature. These tunnels can be accessed by small skin incisions and, in some cases, can be treated with a preferred, one-sided surgical approach [14]. Alambeigi et al. explored the feasibility of using "ortho snake" for bone drilling and bone fixation[15-17]. Their research showed that the "ortho snake" can be used for the drilling of arc channels, which provides the possibility of using an arc channel to fix the pelvic fracture. The arc fixation channel algorithm proposed in this paper provides a basis for robot-assisted arc channel drilling.

In this work, we explore using arc channels for pelvic fracture fixation. We first propose a planning algorithm of pelvic internal arc fixation channel (IAFC) along with an evaluation scheme to measure the feasibility and safety of a fixation channel. Then we generate and evaluate IAFCs on five real pelvis models against straight channels. Our results indicate that IAFCs have higher adaptability compared to traditional straight channels.

## II. MATERIALS AND METHODS

### A. Data acquisition and preprocessing

In order to verify the effectiveness of our proposed arc channel algorithm, our experiment is based on CT models of the pelvic area in five patients with the intact pelvis from Beijing Jishuitan Hospital. The pelvis data is imported into the software Mimics, and the pelvis regions are separated using image processing algorithms which includes threshold segmentation, region growth, open operation, and closed operation, and the separated pelvis model is exported to STL format. The Geomagic studio 12 is used to simplify and smoothen the pelvis model. An isolated pelvis model derived for this study is presented in Fig. 2.

### B. Algorithm design of IAFC

While planning the operation for pelvis fracture fixation, it is often desired that doctors find an optimal fixation channel that has the least possibility of causing the patient harm. We aim to derive such channel using an algorithm. In clinical practice, it is difficult to drill curved channels in human bones. Cable-driven Dexterous Continuum Manipulators (DCM) provide a possibility, but current DCMs can not reach stiffness needed for osseous drilling. An alternative possible method of creating curved fixation channels in bone is to use a tool built with constant curvature: attaching a drill to a rigid metal ring and gradually insert the tube through the target bone. The IAFC algorithm we proposed aims to generate constant curvature paths that are suitable for this drilling method.

The algorithm, shown in Fig. 1, first generates candidate fixation paths and then selects an optimal path. To fixate the pelvis fracture, the screw must go through both sides of the pelvis along with the sacrum, which gives us a rough search region for the paths. We generate seed points in the search regions to create candidate paths and then evaluate them with a quality measure that we define.

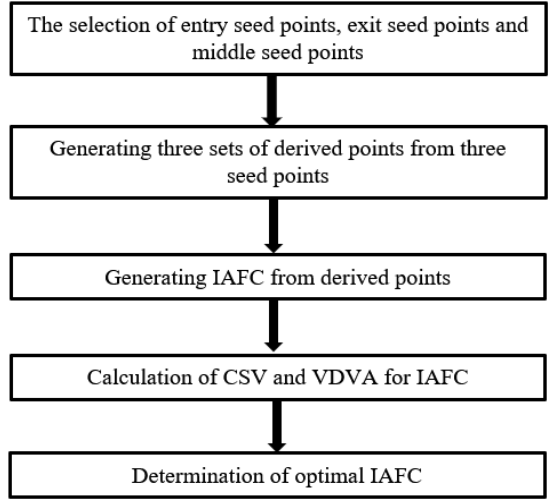

Fig. 1. Overview of the proposed algorithm

(1) IAFC Generation

Three points are needed to determine an arc of constant curvature. We selected two points on both sides of the pelvis as the entry and exit seed points of the IAFC. The entry point and exit point are selected at the upper position of pelvis and hip bone, which is beneficial to improve the stability of screw fixation. A point near S1 on the sagittal plane of the pelvis as the middle seed point of the IAFC is selected, and the computer generates a square lattice as the derived point of the middle seed point. There are two safe areas for screws placed in, among which the front inner side is the safest [18]. According to the position of the seed points, an square lattice is derived near each seed point, which is parallel to the sagittal plane. By projecting a lattice perpendicular to the sagittal plane onto the surface of the pelvis, derivations of the entry and exit points on the surface of the pelvis can be obtained, Fig. 2 shows the generated derived points. During the generation process, points that fall outside the pelvis model will be removed in the algorithm. Taking one point from each of the three sets of points can determine an IAFC with constant curvature. We generate candidate paths using this approach and evaluate them in the next step.

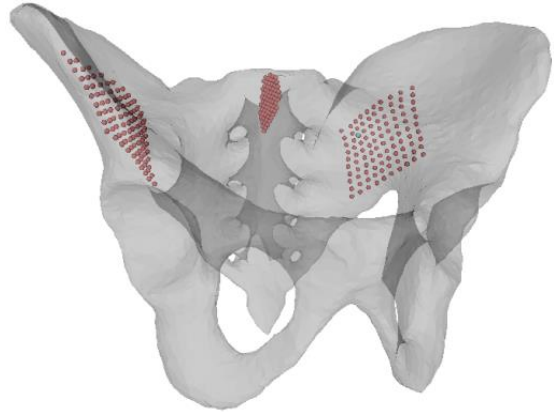

Fig. 2 Derived points on the pelvis. The combination of the points derived from the hip bones of the two groups forms a 10 × 10 lattice. Because of the structural characteristics of S1, the generated lattice is more compact.

(2) Path Safety Evaluation

An evaluation method is necessary to quantify safety of a generated channel and select an optimal arc channel for robot-assisted pelvic fixation. Many of the generated arc channel pass through the pelvis cortex or become close to the pelvic surface, so it is necessary to evaluate their safety. The simplest way to obtain channel safety value (CSV) is to calculate the shortest distance from the centerline of the IAFC to the pelvis surface. However, the combination of the three sets of candidate points will generate an large number of paths, which makes such calculation intractable. We propose a heuristic method to find a balance between realistic safety measure and computational efficiency.

We first discretized a pelvis into a vowel model with voxel size of 1mm × 1mm × 1mm. In order to find the distance from the centerline of a path to the pelvic surface, the pelvis model was corroded by 3D erode algorithm[19]. Every time the algorithm runs, the pelvis' outermost layer of 1mm thickness is removed. The index value of the outermost voxel can be obtained according to the removed position. The voxel distance from these areas to the pelvic surface is 1mm. In the same way, when the erode algorithm runs twice, we can get voxels which are 2mm away from the surface of the pelvis. Therefore, the distance between each voxel and pelvis surface can be calculated by the erode algorithm. These values are defined as the CSV for each voxel. Through the position of the arc channels generated, we can get voxels that these arc channels pass through and the distance safety value from any part of the arc to the pelvis surface inside these voxels.

In order to speed up the calculation, some points with constant intervals on each curve are selected to represent the distance safety value of the whole curve. As shown in Fig. 3, we can get an array of each curve, which we define as voxel distance value array (VDVA). The VDVA of arc channel in the Fig. 3 is [0,1,1,2,2,2,2,2,1,1,0,0]. After the VDVA of each IAFC is obtained, the CSV of the curve can be calculated according to VDVA. The CSV of a fixation path is the least value in the corresponding VDVA.

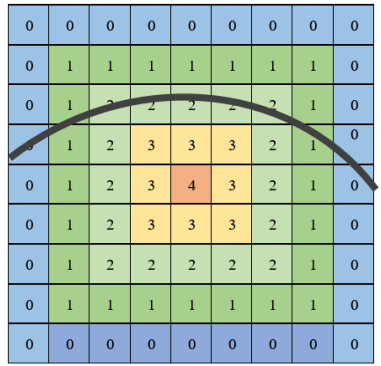

Fig. 3. Two-dimensional schematic diagram of scoring safety values. A two-dimensional square lattice diagram is used to represent a cross section of pelvis to present the erode algorithm. The number in the box indicates the distance of the voxel from the pelvic surface. A voxel of pelvis area is corroded every time until the innermost layer of pelvis.

(3) Selection of the best IAFC

The VDVA of the curve represents the distance from each position on the curve to the pelvis surface, and the CSV of the curve represents the closest distance between the curve and the pelvis surface. The arc channel has a maximum radius, once it exceeds this radius, the channel will pass through the cortical bone. The position of the curve represents the position of the centerline of the constant curvature IAFC. After calculating the corresponding VDVA and CSV of each channel, the channel can be filtered according to these data to select the safest IAFC. We put forward an evaluation method to find an optimal channel. We first find the path with the highest CSV in the set of planned paths, which often results in more than one paths. The second filtering step is to determine the number of minimum values in VDVA and filter the curve with the least number of minimum values in VDVA. Fewer minimum values in VDVA implies less likelihood to reach the bone cortex during drilling of the path. Usually, a singular solution can be obtained after the second filtering. However, if there are still multiple results after the second filtering, the third filtering principle is used to identify the curve with the maximum average value of VDVA.

C. *Calculation of traditional straight sacroiliac screw channels*

Through the above algorithm, a relatively optimal IAFC with constant curvature in the pelvis has been calculated. In order to compare the difference between the traditional straight sacroiliac screw channel and IAFC, a straight channel generation algorithm was obtained based on the proposed IAFC algorithm.

The three sets of points selected in the straight channel generation algorithm are the same as those selected in the arc channel. Similarly, three groups of derived points are generated by using the method of generating derived points of arc channel. The difference between the straight channel and the IAFC algorithm lies in the different curve functions. A straight line is a trivial arc with curvature of zero, so curvature is limited to zero in curve function of the straight channel generation algorithm. Multiple straight channels can be obtained by using the channel generation algorithm. Similarly, by calculating the distance safety value in the voxel through which the straight channel passes, we can also get the VDVA and CSV of each straight channel. An optimal straight channel can be selected from the generated channels by using the method of selecting the best arc channel.

III. Results and Discussion

A. *Comparison of the feasibility of IAFC and straight channel*

The described path planning algorithm is employed to generate both optimal straight (Fig. 4) and optimal arc fixation channels (Fig. 5) on five pelvis models. Despite the individual difference among the pelvis models, the straight optimal paths are rather homogenous, containing similar lengths and tilts. For the IAFCs, however, the difference across the curvature, location, and tilt of the planned paths is much more apparent. It is worth noting that a viable straight path cannot be found on one of the pelvis models due to structural constraints of pelvis. Because the S1 of the pelvis is higher than the position where the pelvis is easy to fix, this structure restricts the penetration of straight channel, so it is impossible to find a straight channel in the pelvis. However, the IAFC is successfully generated. This also preliminarily verities the theory that IAFC is more adaptable and flexible.

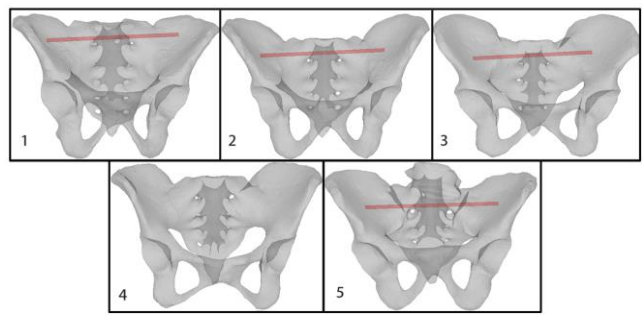

Fig.4 Experimental results of straight channels in each pelvis

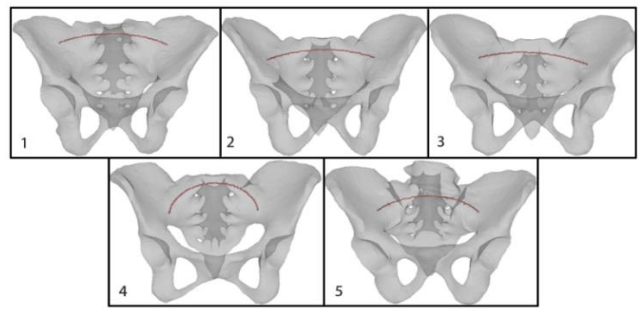

Fig. 5 Experimental results of IAFC in each pelvis

B. *CSV comparison between IAFC and straight channel*

Fig. 6 shows the CSVs of the arc and straight channels in the five pelvis. From the experimental results, we can see that the CSV of IAFC is higher than that of the straight channel. There are many important organs and soft tissues in the pelvis interior [1]. The CSV of straight channel indicates that the position of straight channel is closer to the pelvis surface, which implies that IAFC may reduce the risk of screw implantation better than straight channels.

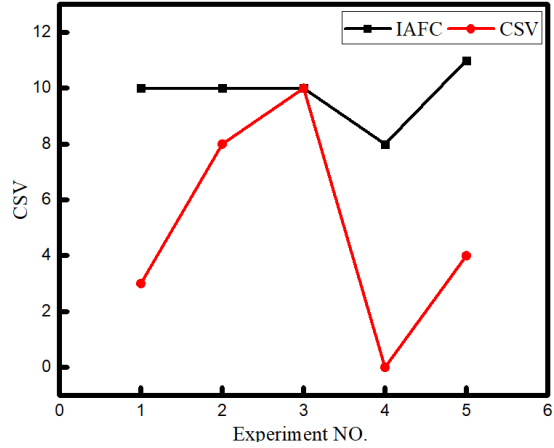

Fig.6 Comparison of IAFC and straight channel CSV

*C. Parameter comparison of IAFC in the different pelvis*

Table I shows the parameters of IAFC in the different pelvis, and the statistical results. As a straight channel is a trivial case of an arc channel, they must be also generated during the IAFC candidate path generation phase. The optimal paths selected by the proposed safety measure all contain varying degrees of stiffness. This indicates that the IAFC are superior to straight channels under the proposed safety measure.

TABLE I. Comparison of parameters of IAFC in the different pelvis

| NO. | Length(mm) | Curvature | CSV |
|---|---|---|---|
| 1 | 162.82 | 0.0089 | 10 |
| 2 | 146.81 | 0.0078 | 10 |
| 3 | 141.56 | 0.0053 | 10 |
| 4 | 154.86 | 0.0187 | 8 |
| 5 | 150.98 | 0.0126 | 11 |

IV. CONCLUSION

In this study, IAFCs in the pelvis are calculated and evaluated. Among the five pelvis models chosen, the IAFCs exhibit higher adaptability and safety than the traditional straight channels. To efficiently gauge the planned fixation paths, we developed a heuristic that quantifies the distance of a path to the pelvis surface, which is often related to the quality of a channel. This approach trades some accuracy of the final safety measure for fast evaluation. Our work aligns with previous studies by Zakarizee et al. [14] in that we both show the viability of IAFCs. But despite the positive theoretical evaluations of IAFCS, many technical challenges need to be overcame before practical implementation. The most significant challenge of which is to stably drill curved paths in the pelvis. Continuum manipulators with variable stiffness, such as [16] and [17], provide possible solutions. If continuum manipulator drills become available in the future, curved pelvis fixation paths with variable curvature may become feasible. Therefore, exploring algorithms for generating variable-curvature pelvis fixation channels may be an extension of this paper. The current algorithm also requires manual selection of initial entry and exit points on the pelvis, making this process autonomous is also possible future work from this study.

# Acknowledgment

All authors declare that they have no conflict of interest.